\documentclass{bmvc2k}

\usepackage{anyfontsize}
\usepackage[10pt]{moresize}
\usepackage{color}
\usepackage{soul}
\usepackage {amsmath}
\usepackage{amssymb}
\usepackage{multicol}
\usepackage{lipsum}
\usepackage{mwe}
\usepackage{indentfirst}
\usepackage{booktabs}
\usepackage{graphicx}
\usepackage[table,xcdraw]{xcolor}
\usepackage{setspace}
\usepackage{times}
\usepackage{hhline}
\usepackage{epsfig}
\usepackage{graphicx}
\usepackage{enumitem}
\usepackage{multirow}
\usepackage{hyperref}
\usepackage{float}

\newcommand\blfootnote[1]{%
	\begingroup
	\renewcommand\thefootnote{}\footnote{#1}%
	\addtocounter{footnote}{-1}%
	\endgroup
}

\title{Gated Multiple Feedback Network for Image Super-Resolution}

\addauthor{Qilei Li\textsuperscript{*}}{qilei.li@outlook.com}{1}
\addauthor{Zhen Li\textsuperscript{*}}{zhenli1031@gmail.com}{1}
\addauthor{Lu Lu}{lulu19900303@126.com}{1}
\addauthor{Gwanggil Jeon}{ggjeon@gmail.com}{23}
\addauthor{Kai Liu}{kailiu@scu.edu.cn}{4}
\addauthor{Xiaomin Yang\textsuperscript{$\dagger$}}{arielyang@scu.edu.cn}{1}

\addinstitution{
	College of Electronics and Information Engineering\\
	Sichuan University\\
	Chengdu, China
}
\addinstitution{
	School of Electronic Engineering\\
	Xidian University\\
	Xi\rq an, China
}
\addinstitution{
	Department of Embedded Systems Engineering\\
	Incheon National University\\
	Incheon, Korea
}
\addinstitution{
	College of Electrical Engineering\\
	Sichuan University\\
	Chengdu, China
}

\runninghead{Li, Li, Lu, Jeon, Liu, Yang}{GMFN for Image Super-Resolution}

\def\eg{\emph{e.g}\bmvaOneDot}

\begin{document}
	
	\maketitle
	
\blfootnote{*Both authors have equally contribution.}
\blfootnote{$\dagger$Corresponding author}	
		\vspace{-\topsep}
	\begin{abstract}
		The rapid development of deep learning (DL) has driven single image super-resolution (SR) into a new era. However, in most existing DL based image SR networks, the information flows are solely feedforward, and the high-level features cannot be fully explored. In this paper, we propose the \textit{gated multiple feedback network} (GMFN) for accurate image SR, in which the representation of low-level features are efficiently enriched by rerouting multiple high-level features. We cascade multiple residual dense blocks (RDBs) and recurrently unfolds them across time. The multiple feedback connections between two adjacent time steps in the proposed GMFN exploits multiple high-level features captured under large receptive fields to refine the low-level features lacking enough contextual information. The elaborately designed gated feedback module (GFM) efficiently selects and further enhances useful information from multiple rerouted high-level features, and then refine the low-level features with the enhanced high-level information. Extensive experiments demonstrate the superiority of our proposed GMFN against state-of-the-art SR methods in terms of both quantitative metrics and visual quality. Code is available at \url{https://github.com/liqilei/GMFN}.
	\end{abstract}

	\section{Introduction}
	\label{sec:intro}
	\setlength{\parskip}{0.5\baselineskip}
	Single image super-resolution (SR) aims to reconstruct a high-resolution (HR) image from its corrupted low-resolution (LR) measurement. It's an ill-posed problem since an LR image can be degraded from multiple HR images. In recent years, the development of deep learning (DL) based high-level vision (skip connections \cite{he2016deep, huang2017densely} and attention mechanism \cite{hu2018squeeze}) helps networks for image SR become much deeper: from 3 layers in SRCNN \cite{dong2014learning} to about 400 layers in RCAN \cite{zhang2018image}, and also made the effects of image SR a truly breakthrough \cite{kim2016accurate, dong2016accelerating, lim2017enhanced, tong2017image, haris2018deep, zhang2018residual}. Nevertheless, as the network deepens, the required parameters are rapidly increasing. To alleviate this problem, the recurrent structures were exploited in \cite{kim2016deeply, tai2017image, han2018image,li2019srfbn}.
	
	However, nearly all the DL based image SR networks are wholly feedforward: the features solely flow from the shallower layers to deeper ones, subsequently, the high-level features extracted from the top layer are directly used to reconstruct an SR image. For these feedforward networks, since the receptive fields in shallower layers are smaller than deeper ones, shallower layers cannot take the valuable contextual information into account. Such a shortcoming hinders the reconstruction ability to some extent. 

	\begin{figure}[tbp]
	\centering
	\includegraphics[width=\textwidth]{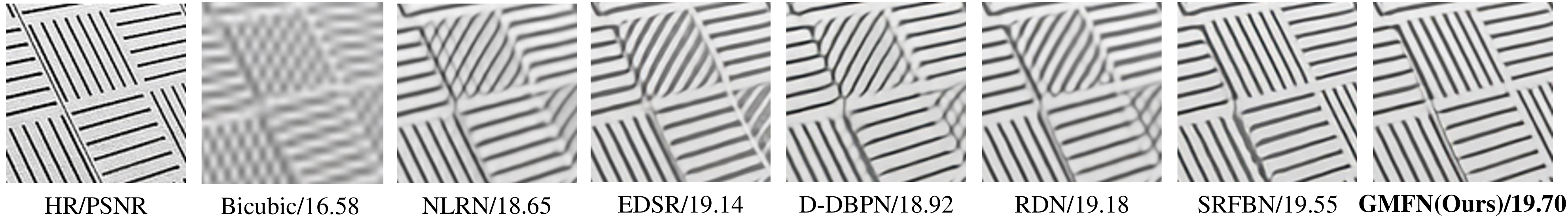}
	\vspace{-5mm}
	\caption{Qualitative results for $\times 4$ image SR on `img\_092' from Urban100 dataset. The proposed GMFN accurately recovers more image details compared with other state-of-the-art image SR methods.}
	\label{fig:vis1}
	\vspace{-6mm}
	\end{figure} 
	
	The feedback mechanism in deep networks aims to refine the low-level features by propagating high-level features to the shallow layers. With the help of high-level information, low-level features become more representative and informative. It has been widely exploited in many high-level vision tasks \cite{carreira2016human, zamir2017feedback,jin2017multi, sam2018top, zhang2018progressive, li2019srfbn} but has rarely been employed for image SR. Although SRFBN \cite{li2019srfbn} explored the feasibility of feedback mechanism for image SR, its feedback connections only propagate the highest-level feature to a shallow layer, other high-level information captured under different sizes of receptive fields is omitted. Hence, such a design neither fully exploits high-level features, nor adequately refines the low-level features.
	
	Based on the above considerations, we propose the gated multiple feedback network (GMFN) for image SR. Since not only the highest-level feature is effective in refining low-level features, we employ multiple feedback connections to transmit multiple high-level features to shallow layers. However, excessive high-level features may be overly redundant, and directly using them may conflict with the original low-level features. Consequently, we design gated feedback modules (GFMs) to adaptively select as well as enhance useful high-level information to refine low-level features. Thanks to the valuable contextual information from the high-level features, the low-level features become more representative, which will intrinsically improve the reconstruction ability of the network. As shown in Fig.~\ref{fig:vis1}, our proposed GMFN shows better visual quality in comparison with other state-of-the-art image SR methods.

	The contributions of our work are summarized as follows:
	\begin{itemize}
		\vspace{-\topsep}
		\setlength{\itemsep}{0pt}
		
		\item We propose the gated multiple feedback network (GMFN) for accurate image SR. Extensive experiments demonstrate the superiority of the proposed GMFN among other state-of-the-art SR methods. Particularly, our final model unfolded with two time steps and each contains 7 residual dense blocks (RDBs) outperforms RDN\cite{zhang2018residual} which employs 16 RDBs.
		
		\item We design the multiple feedback connections to propagate multiple hierarchical high-level features for refining the low-level features. Since high-level features are captured under large receptive fields, they possess more contextual information which is lacking in low-level features. With the help of valuable contextual information introduced by multiple feedback connections, low-level features become more representative, and then the reconstruction performance is intrinsically improved. 
		
		\item We design the simple yet efficient gated feedback module (GFM) to adaptively select and further enhance useful information from multiple rerouted high-level features for refining low-level features. Since only the useful information is permitted to pass, the redundant information among high-level features is efficiently eliminated. The selected and enhanced high-level information enables low-level features to be more informative. 
		
	\end{itemize}
	\vspace{-\topsep}
			\vspace{-4mm}
	\section{Related Work}
	\label{sec:related_work}
	\setlength{\parskip}{0\baselineskip}
	\subsection{Feedback mechanism}
	\label{sec:feedback_mechanism}
	The feedback mechanism in deep network empowers the low-level features to become more representative and informative by propagating the high-level information extracted from deep layers to shallow layers. It has been widely studied for various computer vision tasks (\eg classification~\cite{zamir2017feedback}, pose estimation~\cite{carreira2016human}, and so on~\cite{pinherio2014recurrent, liang2015convolutional, jin2017multi, sam2018top, zhang2018progressive, li2019srfbn}). The majority of feedback connections in these networks are single-to-single, which means only the highest-level features are transmitted to the shallowest layer. Following a different direction, \cite{jin2017multi, zhang2018progressive} applied the single-to-multiple feedback connections to scene parsing and salient object detection, in which the highest-level features are transmitted to multiple shallow layers. They claimed that delivering other high-level features back would introduce redundant information and it might hurt performance on high-level vision tasks. Nevertheless, we argue that such single-to-single and single-to-multiple feedback connection designs are not suitable for image SR task since different levels of features are captured under different receptive fields, every piece of them is significant in reconstructing an SR image.
	
	Taking flaws of previous works into consideration, we introduce new types of feedback connections for accurate image SR, in which multiple hierarchical high-level features are transmitted to shallower layer(s). In other words, the proposed feedback connections are naturally multiple-to-single and multiple-to-multiple. Moreover, we design the gated feedback module to adaptively eliminate redundant information among propagated high-level features, and refine low-level features by using selected high-level information. The valuable contextual knowledge from the high-level information enables the low-level features to be more informative and representative, hence the reconstruction performance is intrinsically improved. Experimental results demonstrate that our gated multiple feedback connections obviously outperform both single-to-single and single-to-multiple ones (see Sec.~\ref{sec:sgmc}).
	
	\subsection{Deep learning based image SR}

	Recently, deep learning based image SR technology has been rapidly developed over the pioneering work \cite{dong2014learning}. The input of the network has changed from the interpolated LR image \cite{dong2014learning,kim2016accurate,kim2016deeply} to the original LR image \cite{ledig2017photo,lim2017enhanced,zhang2018residual}. By doing so, the required computational cost was quadratically saved, and the motion effect caused by interpolation operation was efficiently alleviated. Furthermore, the applications of various skip connections helped the networks went deeper and obtained better reconstruction performance. EDSR \cite{lim2017enhanced} and RCAN \cite{zhang2018image} employed residual skip connections \cite{he2016deep}, SRDenseNet \cite{tong2017image} applied dense skip connections \cite{huang2017densely}, and RDN \cite{zhang2018residual} further integrated residual and dense skip connections together. However, these networks require a huge amount of parameters. 
	
	Recurrent structure can effectively reduce the parameters of the network. It has been widely applied to image SR. Specifically, DRCN \cite{kim2016deeply} and DRRN \cite{tai2017image} can be explained as recurrent neural networks (RNNs) if we regard the input LR image as the initial hidden state, and zero input as the input state \cite{han2018image}. Based on this view, DSRN \cite{han2018image} and NLRN \cite{nlrn2018} designed dual-state and introduced non-local operations for image SR. In these RNN-based methods, however, the information flows from LR image to HR image are solely feedforward. Though SRFBN \cite{li2019srfbn} explored the feasibility of feedback mechanism for image SR by designing an RNN with single-to-single feedback connections which deliver the highest-level features to a shallow layer. We argue that SRFBN fails to fully use other high-level features captured under large receptive fields, thus it cannot efficiently refine the low-level features. 
	
	In contrast, we propose GMFN to make the use of multiple high-level features to enrich the representation of low-level features by recurrently using feedback connections. Except multiple feedback information flows, the proposed GMFN has three other main differences compared with the aforementioned RNN-based methods. First, there are multiple recurrent connections, rather than one or two, between two adjacent time steps in our proposed GMFN. Second, in contrast to block-wise recurrent connections, our recurrent connections in the proposed GMFN can bypass multiple blocks, and thus are more flexible. Third, the features carried by recurrent connections in our proposed GMFN are first sent to the gated feedback module (GFM) for selecting meaningful information rather than directly sent to the recurrent block at the next time step.
	
				\vspace{-4.2mm}	
	\section{Gated Multiple Feedback Network for Image SR}
	\subsection{Network framework}
\begin{figure}[htbp]
		\centering
		\includegraphics[width=\textwidth]{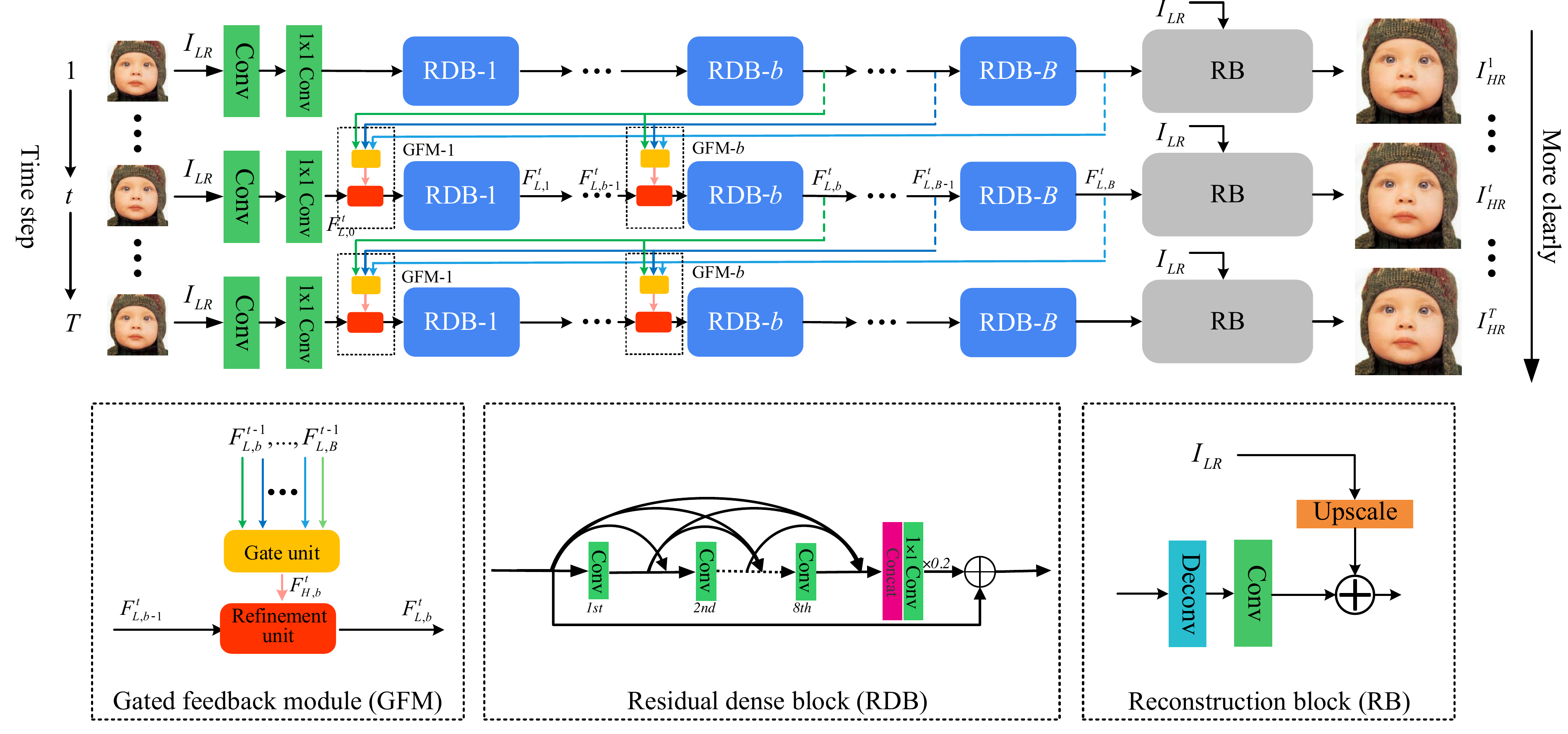}
		\vspace{-5mm}
		\caption{The framework of our proposed gated multiple feedback network (GMFN). }
		\label{fig:gmfn}
		\vspace{-5mm}
	\end{figure}

	As mentioned in \cite{zamir2017feedback, li2019srfbn}, the core merit of a feedback system is propagating output to input in an iterative manner. Following this formulation, our proposed gated multiple feedback network (GMFN) is naturally designed as a convolutional recurrent neural network unrolling $T$ time steps, and the sub-network at each time step can be regarded as an independent convolutional neural network which aims at reconstructing an SR image using an original LR image. As shown in Fig.~\ref{fig:gmfn}, each sub-network mainly consists of four parts: an initial low-level feature extraction block, multiple residual dense blocks (RDBs), multiple gated feedback modules (GFMs), and a reconstruction block (RB). The parameters of these four parts are shared across time. The communication between the sub-networks at two adjacent time steps is achieved by multiple groups of feedback connections. The GFM before one bottom RDB receive one group of feedback connections and further refines the low-level features using selected high-level information.
	
	\label{sec:proposed}
	
	Given $I_{LR}$ as the input image of GMFN at the $t$-th time step, we apply two convolutional layers to extract initial low-level feature $F_{L,0}^{t}$. The first layer and the second layer hold $3\times3$ and $1\times1$ sized convolutional kernels, respectively. $F_{L,0}^{t}$ can be obtained by
	\begin{equation}
	F_{L,0}^{t}=H_{LFEB}\left(I_{LR}\right),
	\end{equation}
	where $H_{LFEB}(\cdot)$ represents the function of the initial low-level feature extraction block. Then the extracted initial low-level feature $F_{L,0}^{t}$ is fed to multiple RDBs to learn hierarchical features. 
	
	Stacking more RDBs will provide more various sizes of receptive field in a sub-network, thus form a better hierarchy of extracted features. 
	Such abundant hierarchical features better assist us in refining low-level features. 
	Each refinement process is accomplished by the GFM placed before one RDB with one group of feedback connections. The details about the GFM will be discussed in Sec.~\ref{sec:gfm}.
	Supposing we cascade $B$ RDBs at each time step, the final high-level feature $F_{L,B}^{t}$ in the LR space can be obtained by
	\begin{equation}
	F_{L,B}^{t}=H_{GFM-RDB}\left(F_{L,0}^{t}\right),
	\end{equation}
	where $H_{GFM-RDB}(\cdot)$ represents the function combining the operations of $B$ RDBs and $M$ GFMs. Specifically, owing to the lack of high-level information provided by the previous time step, there is no GMF placed before any RDB at the first time step (see Fig.~\ref{fig:gmfn}). Following RDN~\cite{zhang2018residual}, the number of the convolutional layers for per RDB is set to 8. 
	
	In reconstruction block, the extracted high-level feature $F_{L,B}^{t}$ is first upscaled by a deconvolutional layer. Then, a $3\times3$ sized convolutional layer recovers a residual image using the upscaled feature. Finally, the recovered residual image is combined with the interpolated LR image to reconstruct the SR image $I^t_{SR}$ at the $t$-th time step. The mathematical formulation of the reconstruction block can be expressed as:
	\begin{equation}
	I^t_{SR}=H_{RB}\left(F_{L,B}^{t}, I_{LR}\right)=H_{UF}\left(F_{L,B}^{t}\right)+H_{\uparrow}\left(I_{LR}\right),
	\end{equation}
	where $H_{RB}(\cdot)$, $H_{UF}(\cdot)$ and $H_{\uparrow}(\cdot)$ represent the functions of the reconstruction block, the deconvolutional layer and the convolutional layer, and interpolated kernel, respectively. 
	
	With $T$ time steps unfolded in the proposed GMFN, we can obtain $T$ SR images totally. Similarly, there are $T$ HR images as the reconstruction target of each sub-network. We adopt $L_{1}$ loss function to optimize our GMFN. The loss function can be formulated as:
	\begin{equation}
	\mathcal{L}(\Theta)=\frac{1}{T}\sum_{t=1}^{T}\left \| I^t_{HR}-I^t_{SR} \right \|_{1}
	\end{equation}
	where $\Theta$ represents the parameter set of GMFN, and $I^t_{HR}$ denotes the target HR image at the $t$-th time step.
	
\subsection{Gated feedback module and multiple feedback connections}
\label{sec:gfm}
The gated feedback module (GFM) is employed to utilize multiple high-level features rerouted from the previous time step to refine the low-level feature extracted from shallow layers. As shown in Fig.~\ref{fig:gmfn}, one GFM is composed of a gate unit and a refinement unit. The gate unit adaptively selects and enhances useful high-level information from multiple high-level features. The refinement unit first refines low-level features by using the selected meaningful high-level information, and further sends the refined low-level feature to the following RDB.
The placement of GFM is determined by the level of features to be refined. According to the relative hierarchical relationship among multiple cascaded RDBs, we choose the input of multiple shallow RDBs as the low-level features need to be refined, and the output of multiple deep RDBs as the high-level features to be rerouted. Since the deepest RDBs can extract the most representative information in the LR space which especially facilitates the refinement processes of initial low-level features, we employ multiple groups of feedback connections to deliver multiple high-level features from the deepest RDBs to the shallowest ones. Each group of feedback connections is handled by one GFM. Let's denote $\mathbf{S}_{M}=\left \{1,2,...,M-1,M\right \}$ as the set of selected indexes of the shallowest $M$ RDBs whose input is regarded as low-level features, and $\mathbf{D}_{N}=\left \{N,N+1,...,B-1,B\right \}$ as the set of selected indexes of the deepest $B-N+1$ RDBs whose output is used to refine these low-level features. At the $t$-th time step, the output of the $b$-th RDB $F_{L,b}^{t}$ can be obtained via

\begin{equation}
F_{L,b}^{t}=\left\{
\begin{array}{lcl}
H_{RDB,b}\left(H_{RU,b}\left ( \left [ F_{H,b}^{t},F_{L,b-1}^{t} \right ] \right )\right), & & \text{if $b \in \mathbf{S}_{M}$ and $t>1$},\\
H_{RDB,b}\left(F_{L,b-1}^{t}\right), & & \text{otherwise},
\end{array} \right.
\setlength{\parskip}{0.5\baselineskip}
\label{equ:rdb}
\end{equation}
where $H_{RDB,b}(\cdot)$ and $H_{RU,b}(\cdot)$ represent the functions of the $b$-th RDB and the refinement unit in the $b$-th GFM, respectively, $\left [ F_{H,b}^{t},F_{L,b-1}^{t} \right ]$ refers to the concatenation of $F_{H,b}^{t}$ and $F_{L,b-1}^{t}$, and $F_{H,b}^{t}$ refers to the selected and enhanced high-level information from multiple high-level features which flow into the $b$-th GFM. These high-level features are extracted from the deepest RDBs, and are then carried by one group of feedback connections. Therefore, the selected and enhanced high-level information $F_{H,b}^{t}$ can be given by
\begin{equation}
F_{H,b}^{t}=\left\{
\begin{array}{lcl}
H_{GU,b}\left ( \left [ F_{L,N}^{t-1},...,F_{L,B}^{t-1} \right ] \right ), & & \text{if $b< N$},\\
H_{GU,b}\left ( \left [ F_{L,b}^{t-1},...,F_{L,B}^{t-1} \right ] \right ), & & \text{otherwise},
\end{array} \right.
\label{equ:gate}
\end{equation}
where $H_{GU,b}(\cdot)$ represents the function of the gate unit in the $b$-th GFM. Based on the relative hierarchical relationship among multiple cascaded RDBs, Eq.~\ref{equ:gate} indicates that the $b$-th GFM only receive the output of RDBs whose indexes are equal or larger than $b$ from the previous time step. For parameter and computation efficiency, we employ two $1\times1$ sized convolutional layers as the gate unit and the refinement unit in the $b$-th GFM, respectively. 

 According to Eq.~\ref{equ:rdb} and Eq.~\ref{equ:gate}, the number of GFMs at each time step (except the first time step) and the number of groups of feedback connections between two adjacent time steps are equal to $M$, and the number of feedback connections in each group is determined by the value of $N$. Thus, we can adjust the values of $M$ and $N$ in selected index sets $\mathbf{S}_{M}$ and $\mathbf{D}_{N}$ to control how many low-level features need to be refined and high-level features will be rerouted, respectively. The mentioned feedback connections in Sec.~\ref{sec:feedback_mechanism} are special cases of our feedback formulation. In detail, we can easily set $N=B$ to achieve single-to-single ($M=1$) or single-to-multiple ($M\neq1$) feedback connection(s) which only routes the highest-level feature back to the shallowest RDB(s). However, since we argue that every piece of high-level information captured under different receptive fields is important for reconstructing an SR image, we set $N\neq B$ to achieve multiple-to-single ($M=1$) and multiple-to-multiple ($M\neq1$) feedback manners which fully exploits high-level features to refine the low-level feature(s). 
		\vspace{-\topsep}
\subsection{Implementation details}
\label{sec:id}
We set unfolded time steps as $T=2$\footnote{For more analysis about time steps please refer to supplement material.}, and cascade $B=7$ RDBs in the sub-network at each time step. Following the previous work~\cite{wang2018esrgan}, the residual scale factor for each RDB is set to 0.2. The number of convolutional kernels in the first layer and the last layer of the sub-network is set to $C_{0}$ and $C_{out}$, respectively. Because we mainly focus on the reconstruction of RGB images, $C_{out}$ naturally equals to 3. The number of convolutional kernels in other layers is set to $C$. In the proposed GMFN, all convolutional and deconvolutional layers are followed by a PReLU \cite{he2015delving} activation function except the last convolutional layer of each RDB and the reconstruction block. In the reconstruction block, a bilinear kernel is used to interpolate the LR image. For different upscale factors, the settings for the deconvolutional layer are same as \cite{li2019srfbn}.
	\vspace{-\topsep}
	\section{Experimental Results}
	
	\subsection{Settings}
	
	\setlength{\parskip}{0.5\baselineskip}
	
	\noindent \textbf{Datasets and evaluation metrics.} We use 800 images from DIV2K for training, and augment training images with scaling, rotations, and flips. For testing, we employ five standard benchmark datasets: Set5~\cite{bevilacqua2012low}, Set14~\cite{zeyde2010single}, B100~\cite{martin2001database}, Urban100~\cite{huang2015single}, and Manga109~\cite{matsui2017sketch}. We generate LR images from HR images by using the Matlab function \textit{imresize} with the option \textit{bicubic}. The SR results are evaluated with PSNR and SSIM~\cite{wang2004image} metrics on Y channel (\textit{i.e.}, luminance) of transformed YCbCr space.
	
	\noindent \textbf{Training settings.} For each iteration, 16 RGB LR patches with a size of $48 \times 48$ are fed to the network. The parameters are initialized using the He's method~\cite{he2015delving}. Adam~\cite{kingma2014adam} is employed to optimize the parameters with an initial learning rate of $2\times10^{-4}$. The learning rate is halved for every $2\times10^{5}$ iterations. The model is implemented under Pytorch framework and trained on an NVIDIA 2080Ti GPU.
	
	\subsection{Study of multiple feedback connections and GFM}
	\label{sec:sgmc}
	
	In the following experiments, the number of convolutional kernels $C_{0}$ and $C$ for the first layer and other layers are set to 128 and 32, respectively. Each model is trained under $2\times10^{5}$ iterations and evaluated on Urban100 dataset with scale factor $\times 4$. 
	
	\begin{figure}
		\centering
		\subfigure[\label{fig:ms}]{\includegraphics[width=0.32\textwidth]{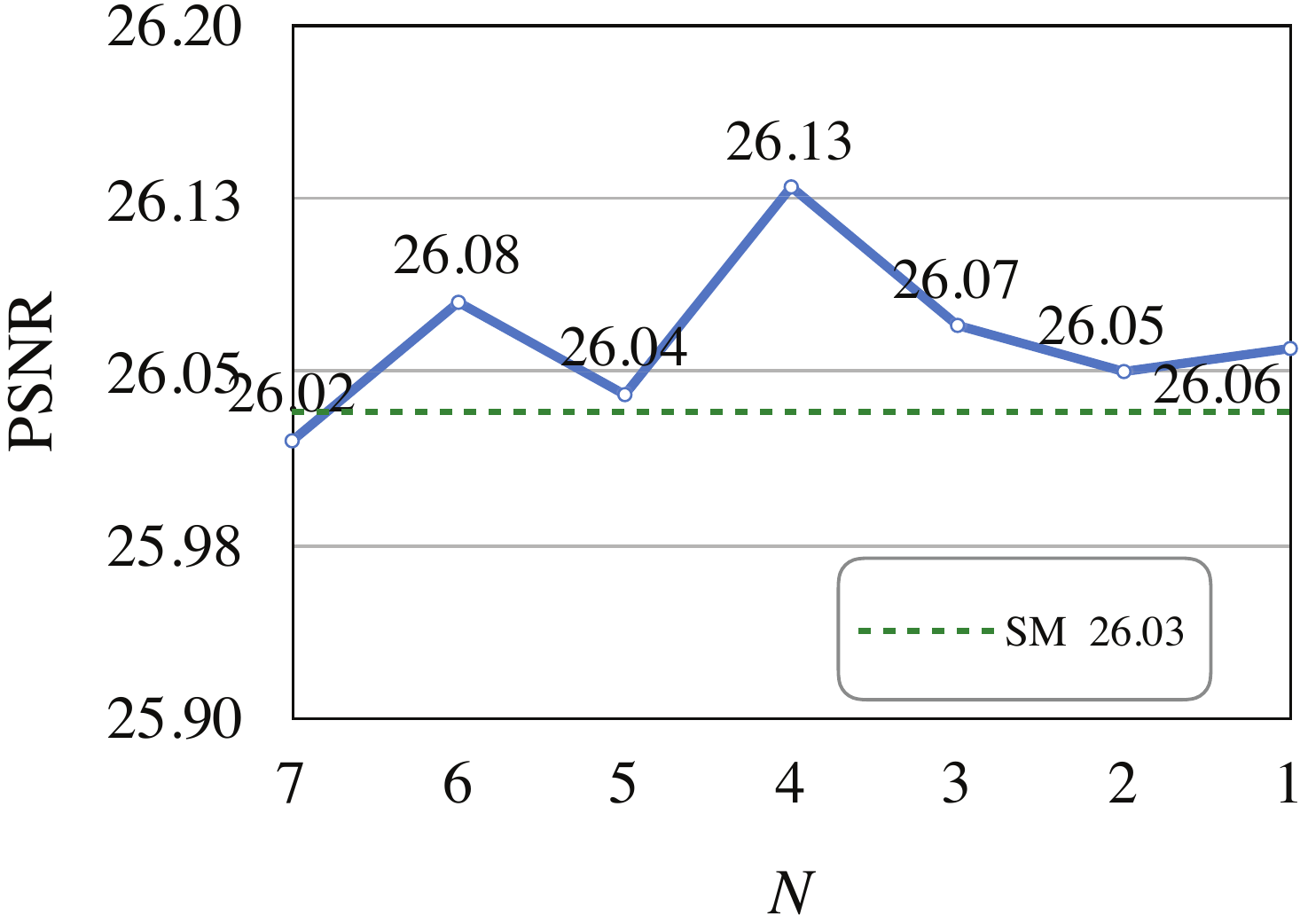}} 
		\subfigure[\label{fig:mm}]{\includegraphics[width=0.32\textwidth]{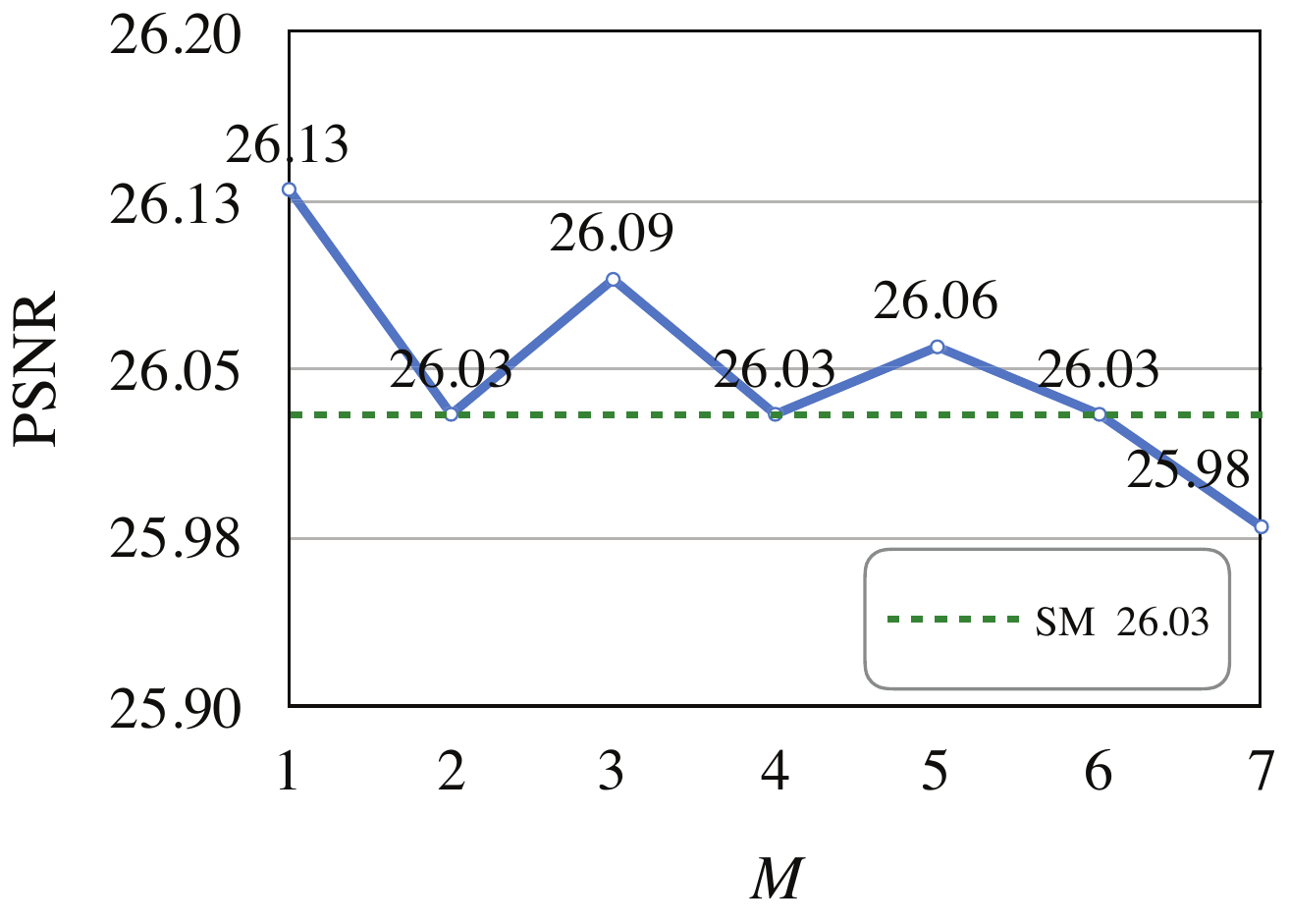}}
		\subfigure[\label{fig:afb}]{\includegraphics[width=0.32\textwidth]{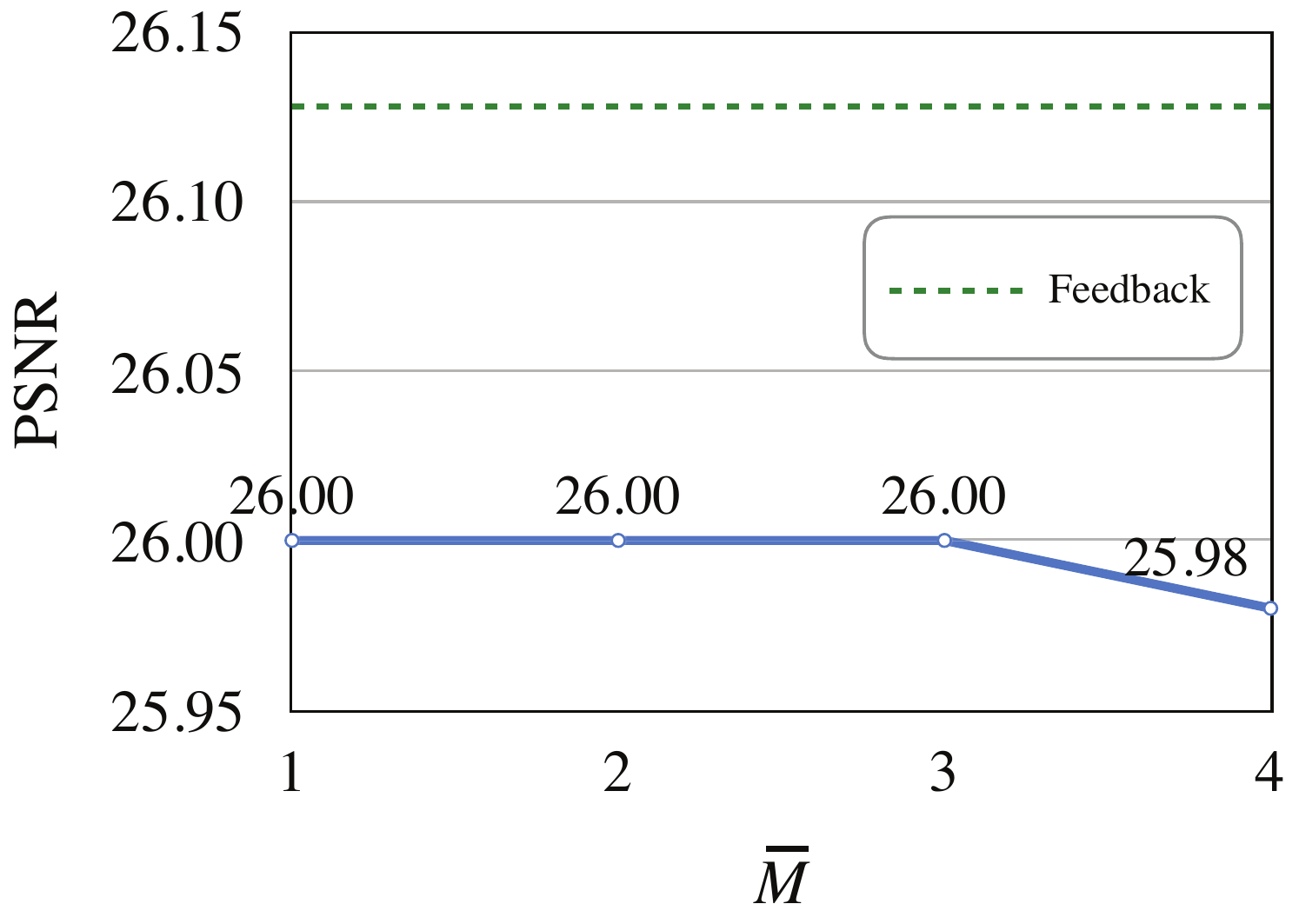}}
		\caption{Study of multiple feedback connections. Single-to-multiple (SM) feedback manner is provided for a better comparison. (a) Performance of various \textit{multiple-to-single} feedback connections. (b) Performance of various \textit{multiple-to-multiple} feedback connections. (c) Performance of various \textit{single-to-multiple} anti-feedback connections. }
		\label{fig:sgmfc}
		\vspace{-6mm}
	\end{figure}
	
	\noindent \textbf{Study of \textit{multiple-to-single} feedback connections}. Multiple-to-single feedback connections aim to transmit multiple high-level features to the first RDB. We compare seven cases of multiple-to-single feedback manners by setting $M=1$ and $N=7, 6, \cdots, 1$ in selected index sets $\mathbf{S}_{M}$ and $\mathbf{D}_{N}$, respectively. Specifically, for $N=7$ among these cases, the feedback connection is single-to-single as SRFBN \cite{li2019srfbn}. For a better comparison, we employ single-to-multiple feedback manner \cite{jin2017multi, zhang2018progressive} as the baseline. Fig.~\ref{fig:ms} illustrates that all multiple-to-single feedback manners perform better than the single-to-multiple and single-to-single ones. As the propagated high-level information increases, the performance of the network gradually improves. This demonstrates that multiple high-level features are beneficial for refining low-level features. However, excessively introducing high-level features may conflict with the original low-level features, thus propagating more high-level features after the peak (\textit{i.e.} $N=4$) will hurt the reconstruction performance of the network.

	\noindent \textbf{Study of \textit{multiple-to-multiple} feedback connections}. We fix $N=4$ with $M=1, 2, \cdots, 7$ to meet the requirements of the multiple-to-multiple manner. Fig.~\ref{fig:mm} shows that as more shallow RDBs receive high-level information, the performance gradually degrades. This is owing to the first RDB has already made full use of the information from the rerouted high-level features. If the high-level features are propagated to other RDBs, they may conflict with newly refined low-level features and hinder the reconstruction ability of the network. Even so, multiple-to-single feedback connections still perform better than the single-to-multiple one. This further illustrates that not only the highest-level feature but multiple ones help to refine low-level features.
	
	\noindent \textbf{Study of anti-feedback connections}. We design anti-feedback connections to further illustrate the effectiveness of the proposed multiple feedback connections. In detail, we reverse the feedback connections to transmit low-level information extracted from the shallowest RDB(s) to the deepest RDB. Similar to the definition of $M$ and $N$, we take $\overline M$ and $\overline N$ to control how many low-level features to be transmitted and how many high-level features to be refined. As opposed to multiple-to-single feedback connections, we set $\overline N = 7$ and combine various $\overline{M}$ to achieve multiple-to-single anti-feedback connections. As can be seen in Fig. \ref{fig:afb}, the anti-feedback connections shows worse reconstruction effect compared with the proposed multiple feedback connections. This demonstrates that exploiting low-level information to enhance high-level features is less efficient than using abundant high-level information to refine low-level features.

\begin{figure}[htbp]
	\centering
	{\includegraphics[width=\textwidth]{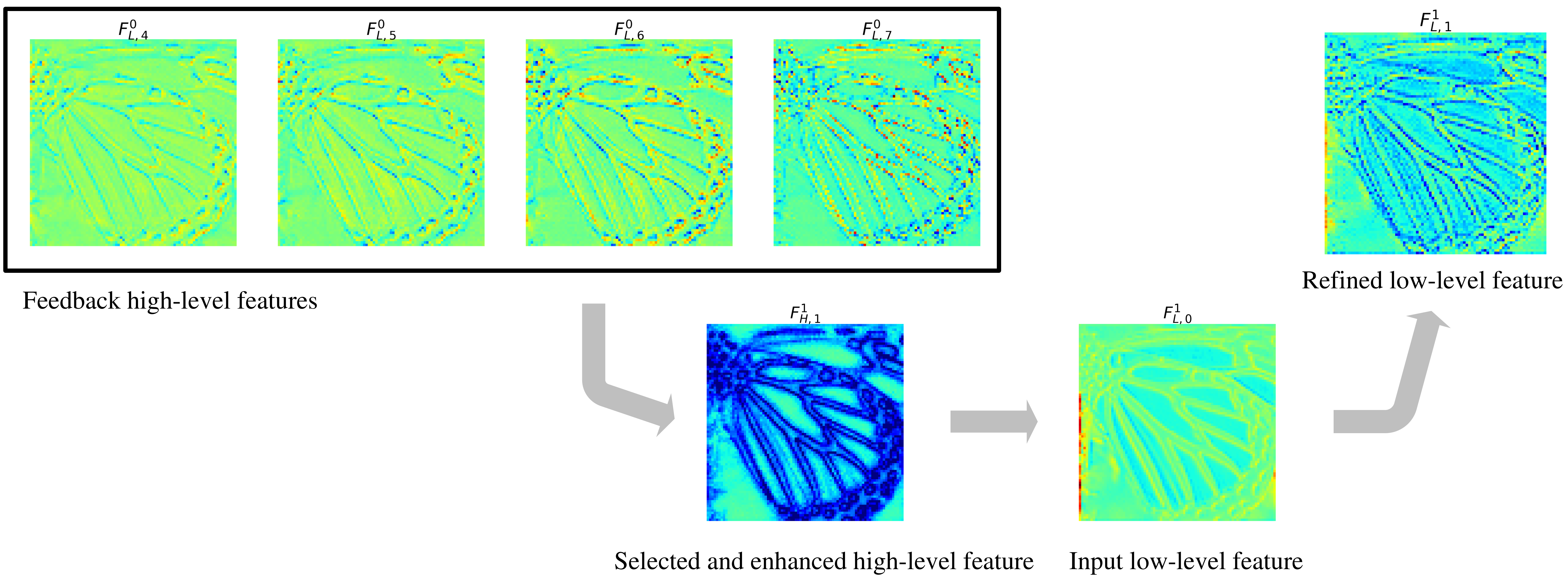}} 
	\vspace{-5mm}
	\caption{Visualization of averaged feature maps.}
	\label{fig:feat}
	\vspace{-4mm}
\end{figure}
	
	\noindent \textbf{Study of the gated feedback module}. The refinement unit in a GFM receives the feedback connection to achieve communication between two adjacent time steps. If we directly remove all GFMs or all refinement units in the GFMs, the communication between the two time step would be disconnected. Thus, we only investigate the necessity of the gate unit in the GFM. For $N=4$ and $M=1$, equipped with the gate unit, our model achieves a PSNR value of 26.13. After removing the gate unit, multiple high-level features are directly concatenated with low-level features at the refinement unit, and the PSNR value under this circumstance drops to 26.06. The reason is that without the gate unit, directly concatenating redundant high-level features with low-level features will confuse the refinement unit, further hinder the reconstruction ability of the network. To better understand the gated feedback module, we visualize the averaged feature maps in Fig. \ref{fig:feat}. As can be seen, the gate unit adaptively selects the high frequency components, such as edges and outlines, of the hierarchical feedback high-level features and generates a more informative high-level features $F_{H,1}^1$. With the help of the selected and enhanced high-level feature, the input low-level feature effectively accesses to high-level information, thus the refined low-level feature becomes more representative than the input low-level feature.

	\subsection{Comparison with the state-of-the-arts}

	\setlength{\parskip}{0\baselineskip}
	
	In this sub-section, the proposed GMFN is equipped with the multiple feedforward connections by setting $M=1$ and $N=4$. $C_{0}$ and $C$ are enlarged to 256 and 64, respectively. We demonstrate the effectiveness of GMFN by comparing it with eight state-of-the-art SR methods: SRCNN~\cite{dong2014learning}, VDSR~\cite{kim2016accurate}, DRRN~\cite{tai2017image}, NLRN~\cite{nlrn2018}, EDSR~\cite{lim2017enhanced}, D-DBPN~\cite{haris2018deep}, RDN~\cite{zhang2018residual}, and SRFBN~\cite{li2019srfbn}. We re-evaluate these comparison methods in accordance with corresponding public implementations and report the quantitative and qualitative comparison results\footnote{Comparison on running time and number of parameters is available in supplement material} in Tab.~\ref{comp_sot_bi} and Fig.~\ref{fig:comp}, respectively. As can be seen, the proposed GMFN performs best on both PSNR and SSIM metrics in most public datasets. Especially, with only 14 RDBs ($T=2$, $B=7$), our GMFN exhibits better reconstruction performance than RDN which has 16 RDBs. The qualitative results shown in Fig.~\ref{fig:comp} indicate our GMFN can reconstruct a faithful SR image with sharper and clearer edges. It can recover more image details compared with other methods. The consistency between quantitative and qualitative results convincingly proves the superiority of the proposed GMFN. 

	\begin{table*}[htbp]
		\centering	
		\resizebox{\textwidth}{!}{\begin{tabular}{|c|c|c|c|c|c|c|c|c|c|c|c|}
			\hline
				\multirow{2}{*}{Dataset} & \multirow{2}{*}{Scale} & \multirow{2}{*}{Bicubic} &          SRCNN          &          VDSR          &        DRRN         &      NLRN       &                 EDSR                 &          D-DBPN          &                 RDN                  &                SRFBN                 &                 GMFN                 \\
			                                        &                        &                          & \cite{dong2014learning} & \cite{kim2016accurate} & \cite{tai2017image} & \cite{nlrn2018} &        \cite{lim2017enhanced}        &   \cite{haris2018deep}   &       \cite{zhang2018residual}       &          \cite{li2019srfbn}          &                (Ours)                \\ \hline\hline
			         \multirow{3}{*}{Set5}          &       $\times2$        &       33.66/0.9299       &      36.66/0.9542       &      37.53/0.9590      &    37.74/0.9591     &  38.08/0.9610   &            {38.11}/0.9602            &       38.09/0.9600       &    \textbf{38.24}/\textbf{0.9614}    &             38.11/0.9609             & \underline{38.21}/\underline{0.9612} \\
			                   -                    &       $\times3$        &       30.39/0.8682       &      32.75/0.9090       &      33.67/0.9210      &    34.03/0.9244     &  34.30/0.9271   &             34.65/0.9280             &           -/-            &  \underline{34.71}/\textbf{0.9296}   &             34.70/0.9292             &  \textbf{34.73}/\underline{0.9295}   \\
			                                        &       $\times4$        &       28.42/0.8104       &      30.48/0.8628       &      31.35/0.8830      &    31.68/0.8888     &  31.94/0.8920   &             32.46/0.8968             & \underline{32.47}/0.8980 & \underline{32.47}/\underline{0.8990} &     \underline {32.47}/{0.8983}      &    \textbf{32.55}/\textbf{0.8991}    \\ \hline\hline
			        \multirow{3}{*}{Set14}          &       $\times2$        &       30.24/0.8688       &      32.45/0.9067       &      33.05/0.9130      &    33.23/0.9136     &  33.57/0.9167   &            {33.92}/0.9195            &       33.85/0.9190       &  \underline{34.01}/\textbf{0.9212}   &            33.82/{0.9196}            &  \textbf{34.05}/\underline{0.9211}   \\
			                                        &       $\times3$        &       27.55/0.7742       &      29.30/0.8215       &      29.78/0.8320      &    29.96/0.8349     &  30.25/0.8386   &           {30.52}/{0.8462}           &           -/-            & \underline{30.57}/\underline{0.8468} &             30.51/0.8461             &    \textbf{30.58}/\textbf{0.8473}    \\
			                                        &       $\times4$        &       26.00/0.7027       &      27.50/0.7513       &      28.02/0.7680      &    28.21/0.7721     &  28.44/0.7759   &       28.80/\underline{0.7876}       & \underline{28.82}/0.7860 &             28.81/0.7871             &             28.81/0.7868             &    \textbf{28.84}/\textbf{0.7888}    \\ \hline\hline
			         \multirow{3}{*}{B100}          &       $\times2$        &       29.56/0.8431       &      31.36/0.8879       &      31.90/0.8960      &    32.05/0.8973     &  32.18/0.8991   & \underline{32.32}/\underline{0.9013} &       32.27/0.9000       &    \textbf{32.34}/\textbf{0.9017}    &             32.29/0.9010             &    \textbf{32.34}/\textbf{0.9017}    \\
			                                        &       $\times3$        &       27.21/0.7385       &      28.41/0.7863       &      28.83/0.7990      &    28.95/0.8004     &  29.05/0.8024   &       {29.25}/\textbf{0.8093}        &           -/-            &  \underline {29.26}/\textbf{0.8093}  &      29.24/\underline {0.8084}       &    \textbf{29.27}/\textbf{0.8093}    \\
			                                        &       $\times4$        &       25.96/0.6675       &      26.90/0.7101       &      27.29/0.7260      &    27.38/0.7284     &  27.48/0.7304   &       27.71/\underline{0.7420}       & \underline{27.72}/0.7400 &      \underline{27.72}/{0.7419}      &       \underline{27.72}/0.7409       &    \textbf{27.74}/\textbf{0.7421}    \\ \hline\hline
			       \multirow{3}{*}{Urban100}        &       $\times2$        &       26.88/0.8403       &      29.50/0.8946       &      30.77/0.9140      &    31.23/0.9188     &  31.77/0.9243   &      \underline{32.93}/{0.9351}      &       32.55/0.9324       &       32.89/\underline{0.9353}       &             32.62/0.9328             &    \textbf{32.96}/\textbf{0.9361}    \\
			                                        &       $\times3$        &       24.46/0.7349       &      26.24/0.7989       &      27.14/0.8290      &    27.53/0.8378     &  27.90/0.8443   & \underline{28.80}/\underline{0.8653} &           -/-            & \underline{28.80}/\underline{0.8653} &             28.73/0.8641             &    \textbf{28.87}/\textbf{0.8667}    \\
			                                        &       $\times4$        &       23.14/0.6577       &      24.52/0.7221       &      25.18/0.7540      &    25.44/0.7638     &  25.78/0.7713   & \underline{26.64}/\underline{0.8033} &       26.38/0.7946       &             26.61/0.8028             &             26.60/0.8015             &   \textbf {26.69}/\textbf{0.8048}    \\ \hline\hline
			       \multirow{3}{*}{Manga109}        &       $\times2$        &       30.30/0.9339       &      35.60/0.9663       &      37.22/0.9750      &    37.60/0.9736     &  38.55/0.9768   &            {39.10}/0.9773            &       38.89/0.9775       &    \textbf{39.18}/\textbf{0.9780}    &       39.08/\underline{0.9779}       &       \underline{39.13}/0.9778       \\
			                                        &       $\times3$        &       26.95/0.8556       &      30.48/0.9117       &      32.01/0.9340      &    32.42/0.9359     &  33.24/0.9414   &             34.17/0.9476             &           -/-            &       34.13/\underline{0.9484}       &       \underline{34.18}/0.9481       &    \textbf{34.24}/\textbf{0.9487}    \\
			                                        &       $\times4$        &       24.89/0.7866       &      27.58/0.8555       &      28.83/0.8870      &    29.18/0.8914     &  29.82/0.8982   &             31.02/0.9148             &       30.91/0.9137       &             31.00/0.9151             & \underline{31.15}/\underline{0.9160} &    \textbf{31.24}/\textbf{0.9174}    \\ \hline
		\end{tabular}}
		\vspace{-3mm}
		\caption{Quantitative evaluation under scale factors $\times 2$, $\times 3$ and $\times 4$. The best performance is shown in \textbf{bold} and the second best performance is \underline{underlined}.
			\label{comp_sot_bi}}
		\vspace{-3mm}

	\end{table*}

	\begin{figure}[H]
		\centering
		\includegraphics[width=\textwidth]{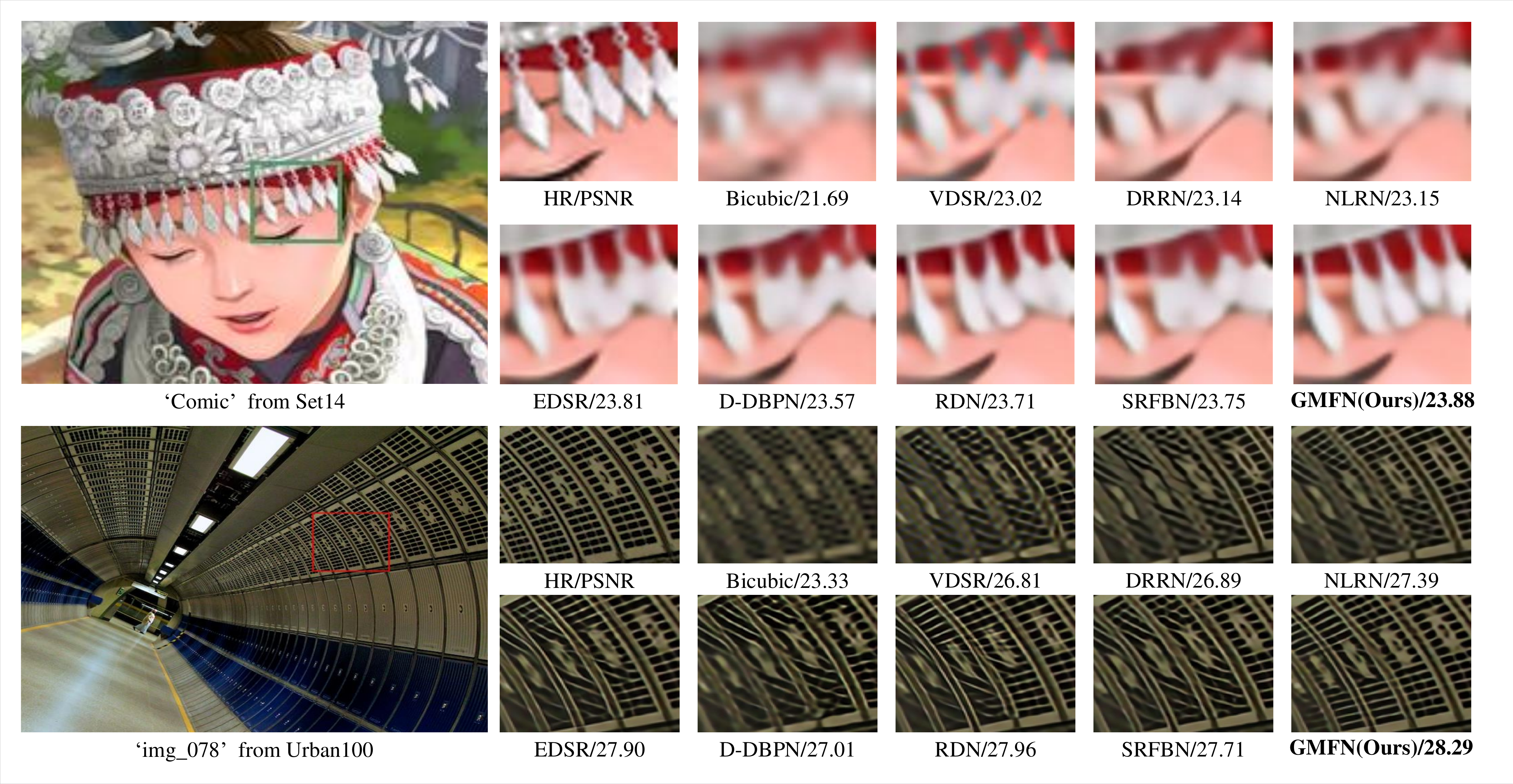}
		\vspace{-5mm}
		\caption{Qualitative comparison of our GMFN with other methods on $\times$4 image SR.}
		\label{fig:comp}
		\vspace{-4mm}
	\end{figure} 	

	\section{Conclusion}	
	In this paper, we propose the gated multiple feedback network (GMFN) for accurate image SR. It successfully enriches the representation of low-level features by propagating multiple hierarchical high-level features to shallow layers. The elaborately designed gated feedback module (GFM) efficiently selects and enhances meaningful high-level information from multiple groups of feedback connections and uses the selected and enhanced high-level information to refine the low-level features. Extensive experiments on investigating and analyzing various feedback manners demonstrate the superiority of our proposed multiple feedback connections. With two time steps and each contains 7 RDBs, the proposed GMFN achieves better reconstruction performance compared to state-of-the-art image SR methods including RDN\cite{zhang2018residual} which contains 16 RDBs.
	
	\vspace{-4mm}
	\section*{Acknowledgement}
	\noindent The research in our paper is sponsored by National Natural Science Foundation of China (No.61701327 and No.61711540303), Science Foundation of Sichuan Science and Technology Department (No.2018GZ0178).
	
	\bibliography{Definitions/egbib}
	
	\clearpage
	\appendix

\begin{center} 
	\noindent\textbf{\Large{Supplementary Material}}
\end{center}
\vspace{4mm}

The following items are contained in the supplementary material:

\ref{sec:ffvsfb}. Feedback networks vs. feedforward networks.

\ref{sec:sot}. Study of time step.

\ref{sec:ma}. Model analysis.

\ref{sec:mqr}. More qualitative results.

\section{Feedback networks vs. feedforward networks}
\label{sec:ffvsfb}

\begin{figure}[htbp]
	\begin{minipage}[t]{0.5\linewidth}
		\centering
		\includegraphics[width=0.7\textwidth]{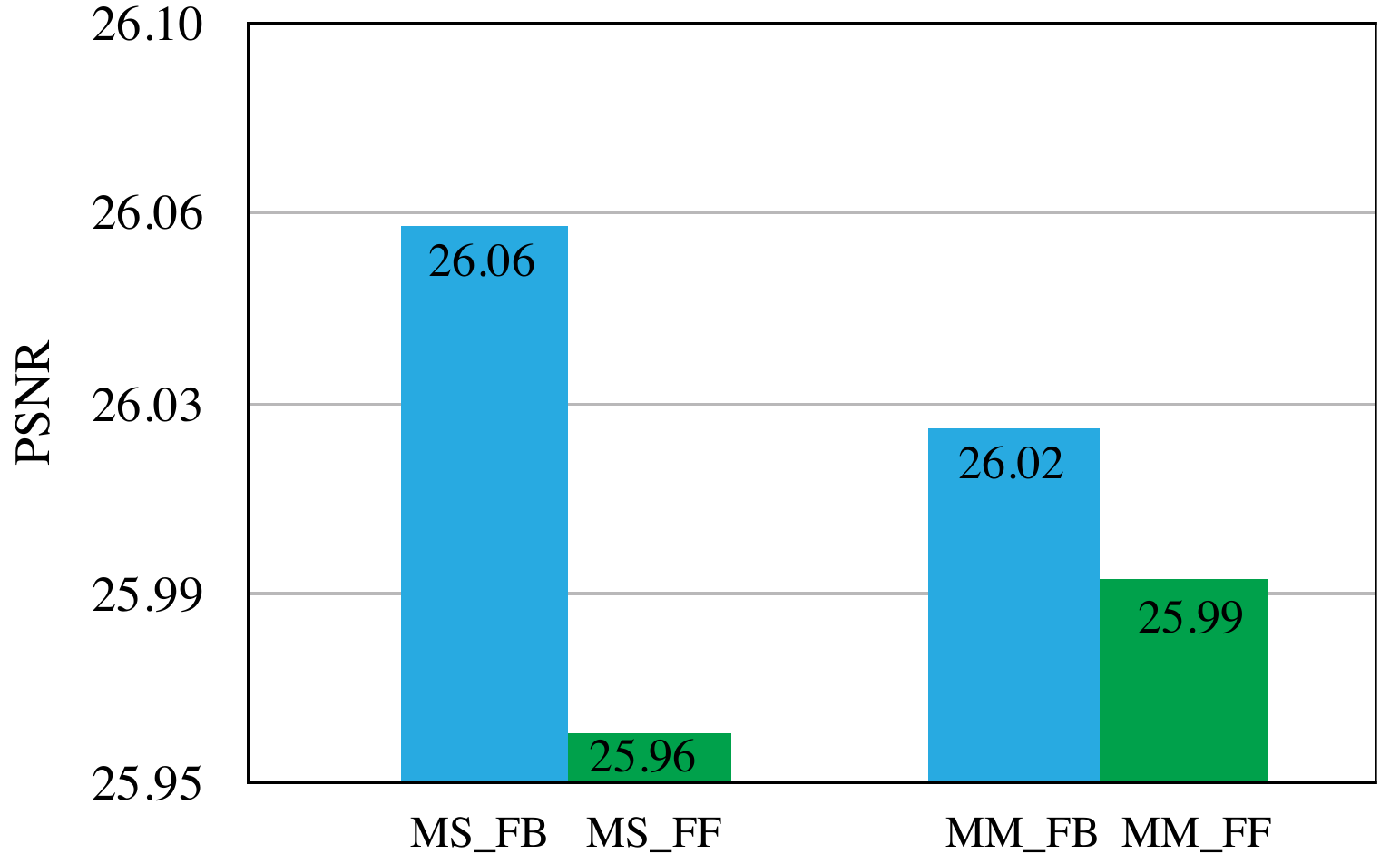}
		\caption{Feedback networks (FB) vs. feedforward networks (FF)}
		\label{fig:ffvsfb}
	\end{minipage}%
	\begin{minipage}[t]{0.5\linewidth}
		\centering
		\includegraphics[width=0.95\textwidth]{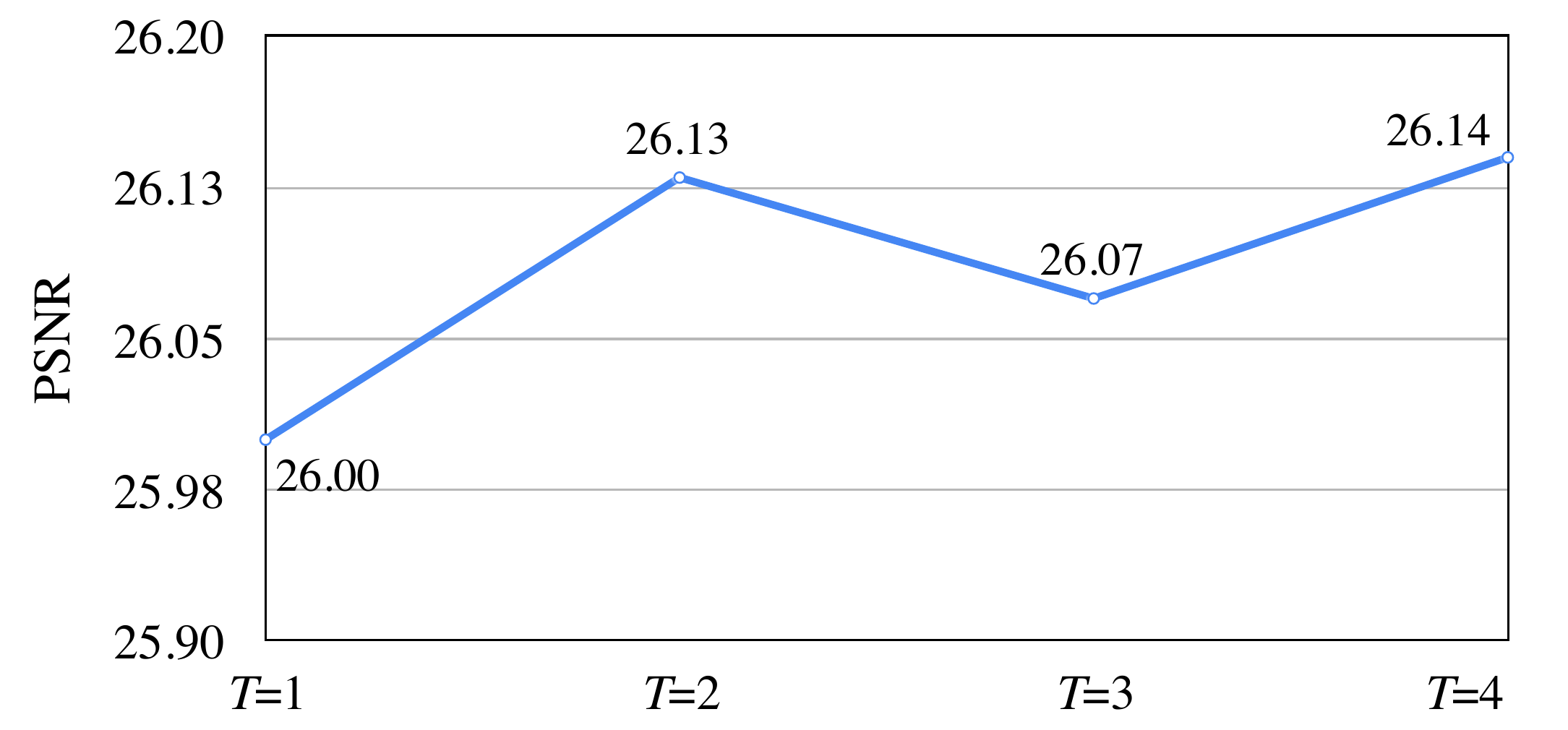}
		\caption{Study of time step (\textit{T})}
		\label{fig:st}
	\end{minipage}
\end{figure}

 We testify the superiority of our feedback networks over the corresponding feedforward networks. The number of convolutional kernels $C_{0}$ and $C$ for the first layer and other layers are set to 128 and 32, respectively. All models are trained under $2\times10^{5}$ iterations and evaluated on Urban100 dataset. We set $M=1$ and $N=1$ to represent \textit{multiple-to-single} (MS) feedback network, $M=7$ and $N=1$ to represent \textit{multiple-to-multiple} (MM) one, and mark them with `FB'. Their feedforward counterparts (marked with `FF') are implemented by disconnecting the loss to all time steps except the last one \cite{zamir2017feedback, li2019srfbn}. The experimental results shown Figure~\ref{fig:ffvsfb} indicate that both MM\_FB and MS\_FB feedback networks outperform the corresponding feedforward networks. This confirms that our gated multiple feedback network has obvious advantages over the traditional feedforward networks. 

\begin{figure}
	\centering
	\subfigure[\label{fig:speed}]{\includegraphics[width=0.43\textwidth]{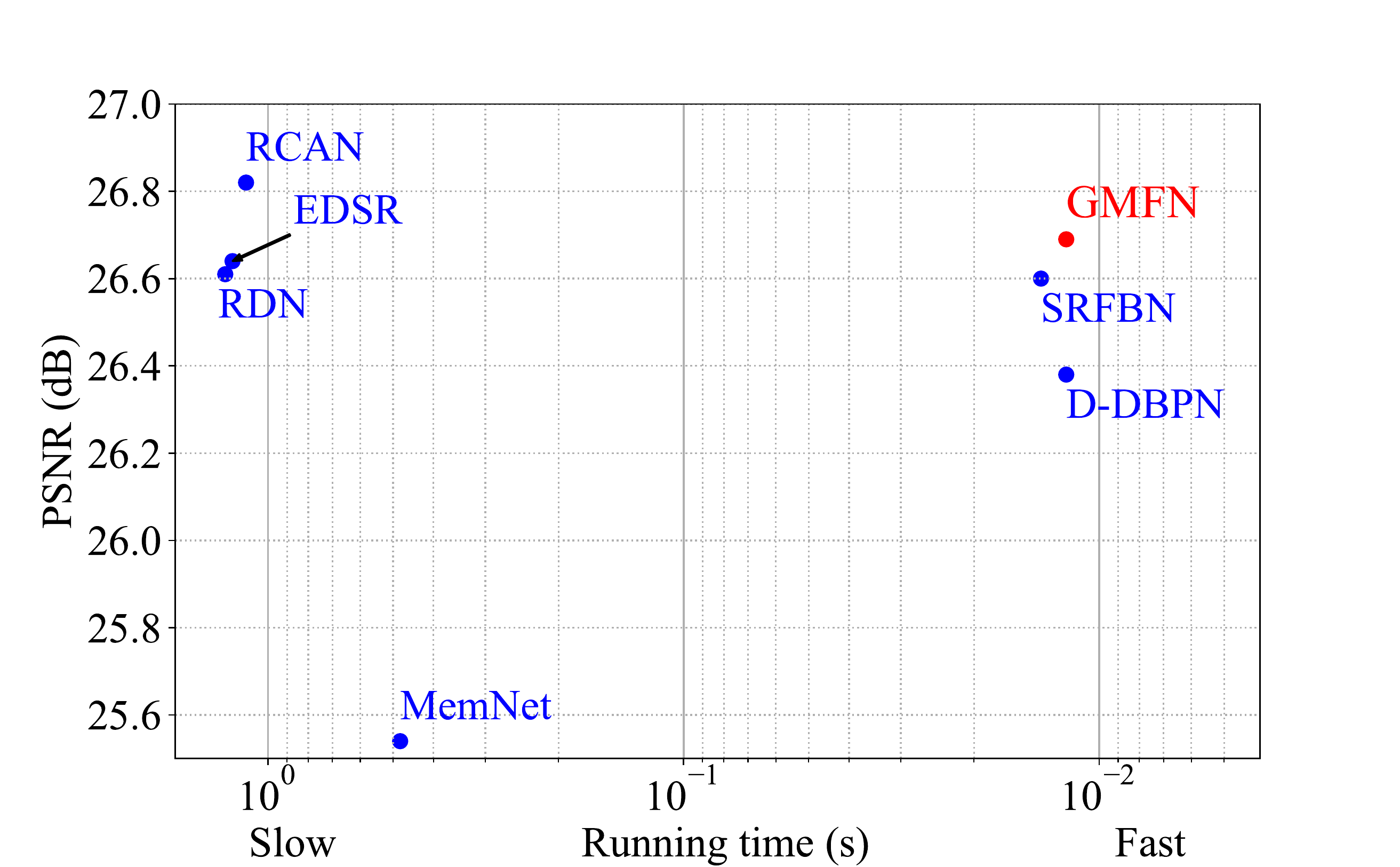}} 
	\subfigure[\label{fig:num_param}]{\includegraphics[width=0.55\textwidth]{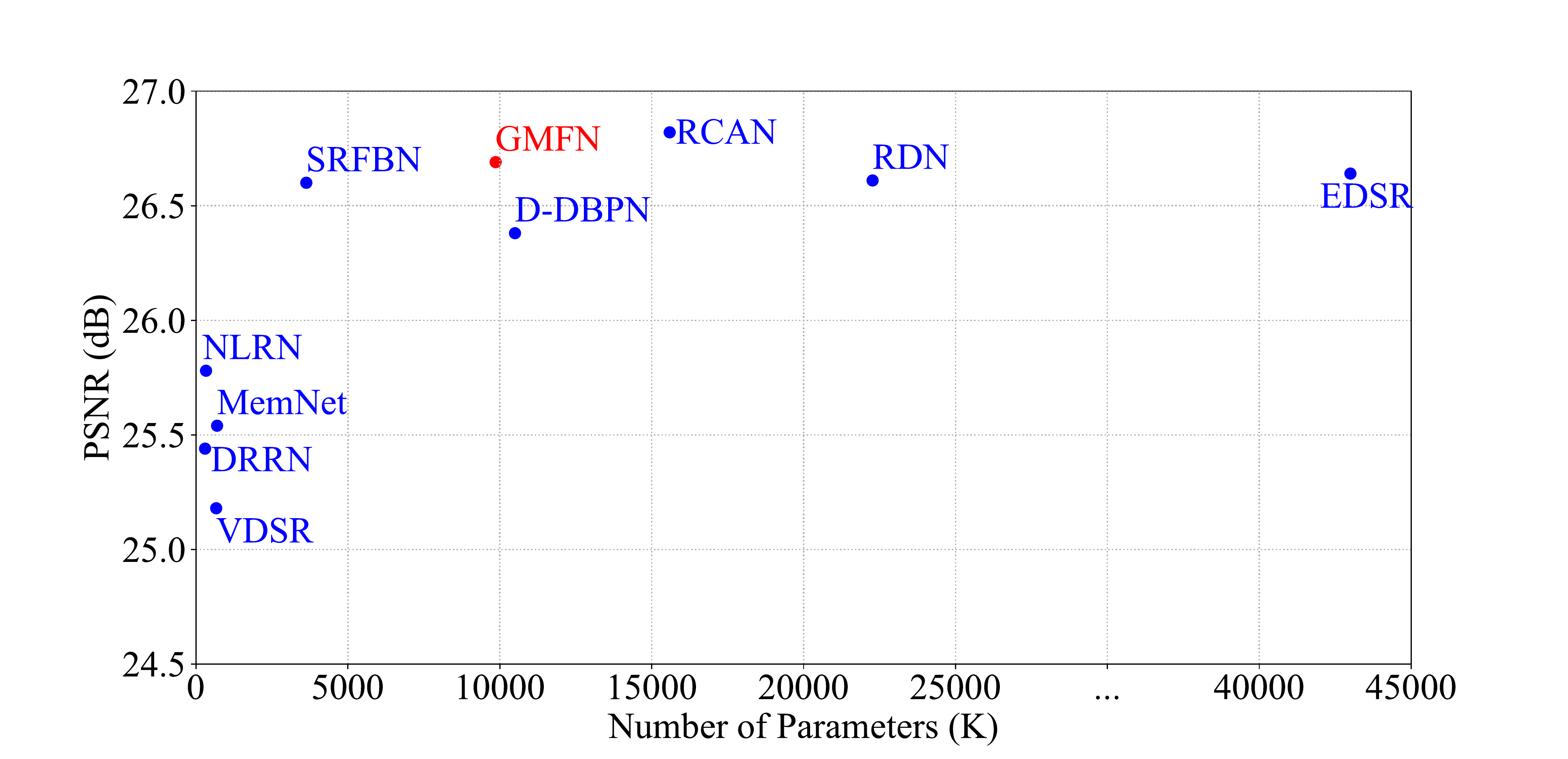}}
	\caption{(a) Accuracy and numbers of parameters trade-off. (b) Accuracy and average running time trade-off. Models are evaluated on Urban100 under scale factor $\times 4$ on an NVIDIA 1080Ti GPU with an i7-7700K CPU.}
	\label{fig:ma}
\end{figure}

\section{Study of time step}
\label{sec:sot}

In this section, we investigate the influence of the time step $T$ on the proposed GMFN. Under identical setting of multiple feedback connection, we set $T$ = 1, 2, 3, and 4, respectively. The performance evaluated on Urban100 dataset is shown in Figure~\ref{fig:st}. It can be observed that with the help of multiple feedback connections, the reconstruction ability is significantly improved compared with the one without feedback connections ($T$=1). However, we also observed as $T$ continues to increase, the reconstruction quality improves slightly. Hence, we set $T=2$ in our main paper for better balance the reconstruction performance and the computational cost. 

\section{Model analysis}
\label{sec:ma}
We compare the running time and the number of parameters of our final model with some representative state-of-the-art methods on Urban100 with scale factor $\times 4$. Figure \ref{fig:ma} shows that the proposed GMFN can well balance the reconstruction accuracy, running time, as well as number of parameters. In terms of running time, GMFN runs orders of magnitude faster than RDN~\cite{zhang2018residual}, EDSR~\cite{lim2017enhanced}, and so on. Compared with SRFBN~\cite{li2019srfbn} and D-DBPN~\cite{haris2018deep}, which require similar running time, GMFN achieves a better reconstruction performance. Additionally, GMFN requires 6\% fewer parameters than D-DBPN, 56\% fewer parameters than RDN, and 77\% fewer parameters than EDSR while obtaining a higher PSNR value. RCAN~\cite{zhang2018image} can attain a better reconstruction performance than all other comparison methods, but it holds relatively more parameters and a much deeper network design (about 400 layers). We will further extend our work following such design. 

\section{More qualitative results}
\label{sec:mqr}
\begin{figure}[htbp]
	\centering
	{\includegraphics[width=\textwidth]{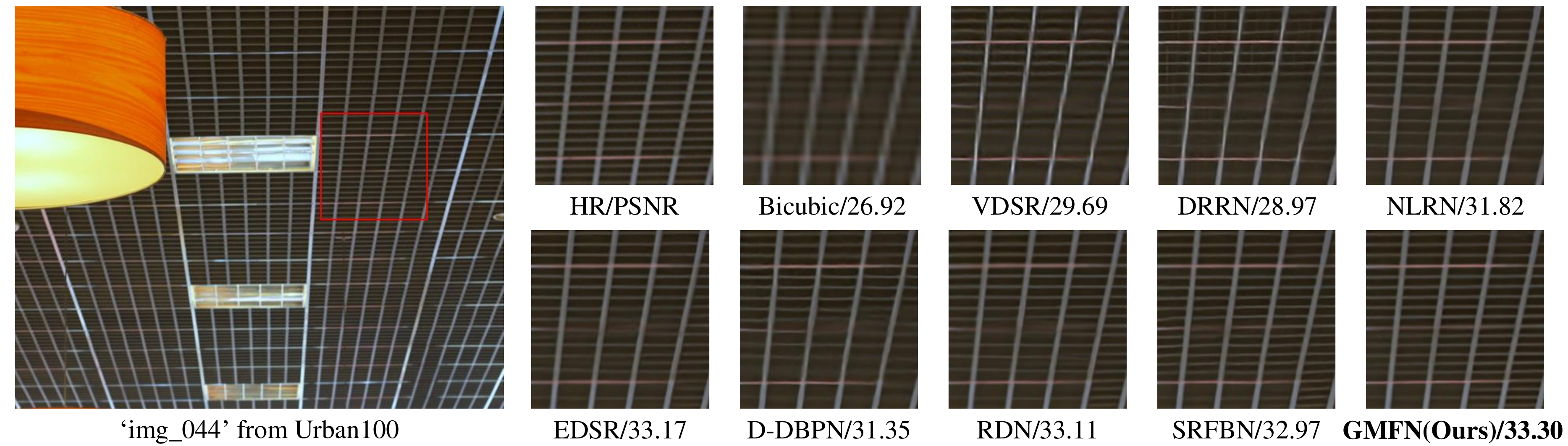}} 
	\caption{Qualitative results on `img\_044' with scale factor $\times$4. The proposed GMFN better recovers grids on the ceiling.}
	\label{fig:img_044}
\end{figure}

\begin{figure}[htbp]
	\centering
	{\includegraphics[width=\textwidth]{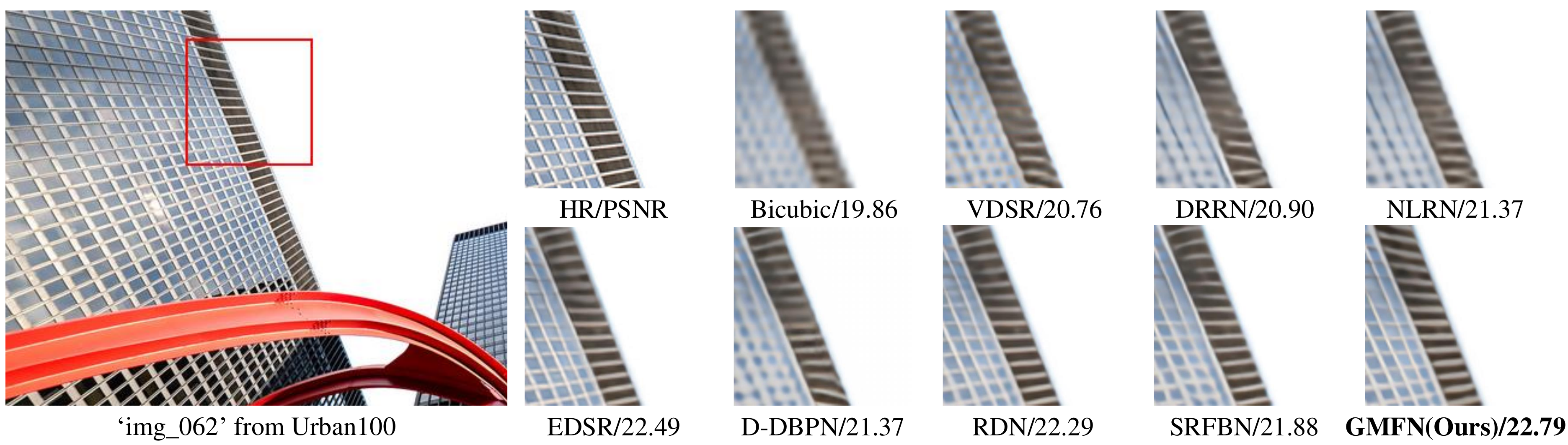}} 
	\caption{Qualitative results on `img\_062' with scale factor $\times$4. The proposed GMFN produces a faithful SR image and avoids artifacts as other methods.}
	\label{fig:img_062}
\end{figure}

\begin{figure}[htbp]
	\centering
	{\includegraphics[width=\textwidth]{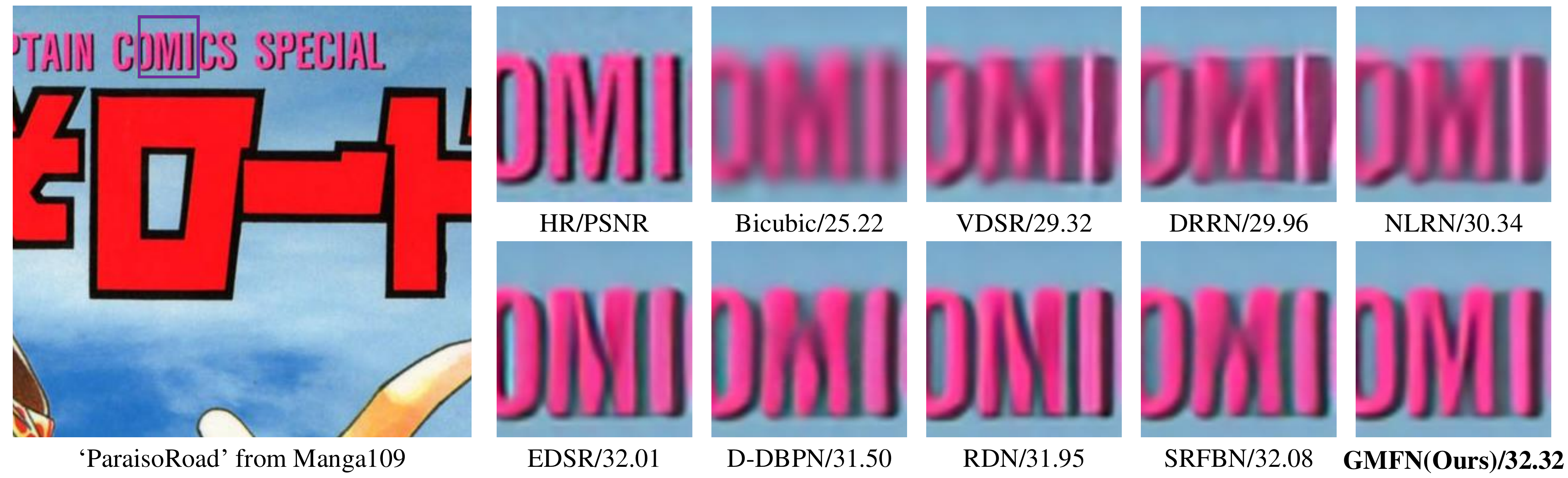}} 
	\caption{Qualitative results on `ParaisoRoad' with scale factor $\times$4. Only the GMFN accurately restored the letter "M", and the results of the other methods broke.}
	\label{fig:ParaisoRoad}
\end{figure}

\begin{figure}[htbp]
	\centering
	{\includegraphics[width=\textwidth]{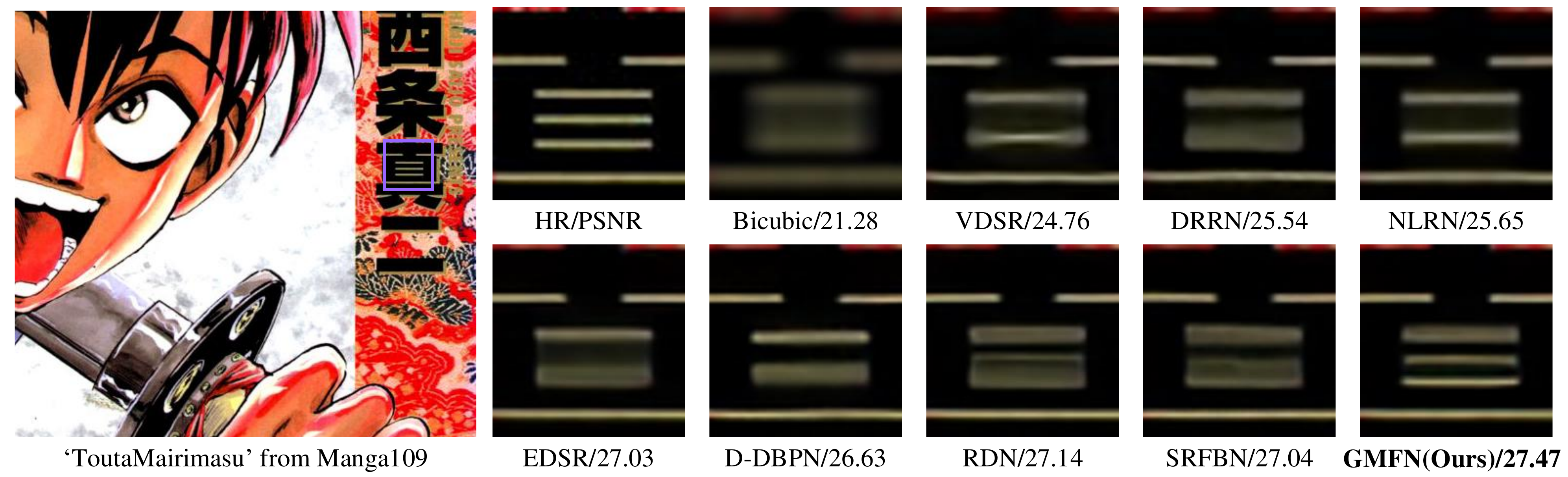}} 
	\caption{Qualitative results on `ToutaMairimasu' with scale factor $\times$4. Only GMFN recoveres two horizontal lines as the HR image. Other comparison methods only restore one horizontal line erroneously.}
	\label{fig:ToutaMairimasu}
\end{figure}

\begin{figure}[htbp]
	\centering
	{\includegraphics[width=\textwidth]{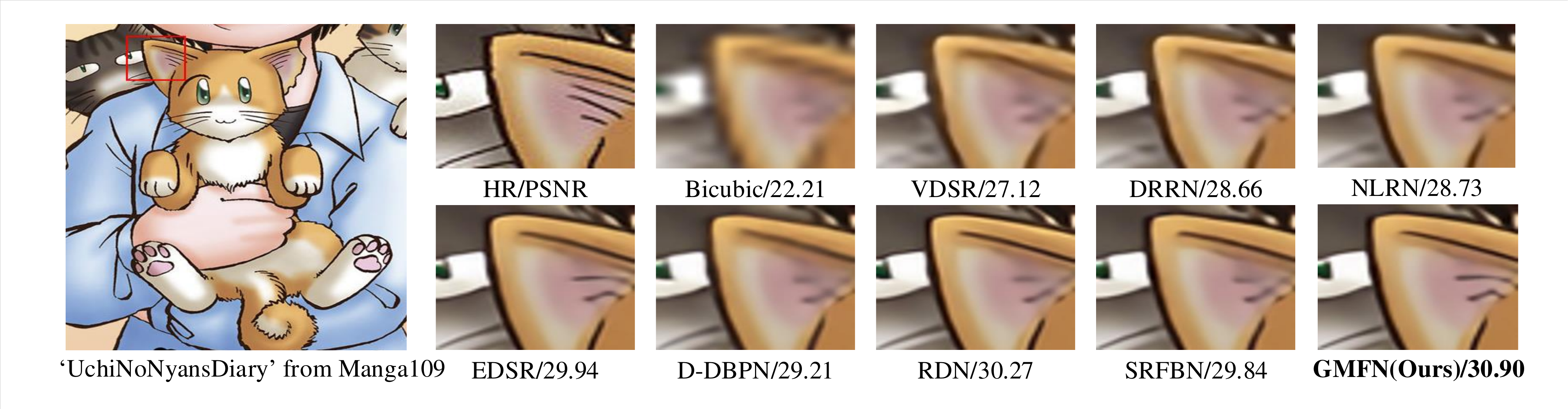}} 
	\caption{Qualitative results on `UchiNoNyansDiary' with scale factor $\times$4. Only GMFN faithful recovers detail of headwear, while other comparison methods cause heavy blurring artifacts.}
	\label{fig:UchiNoNyansDiary}
\end{figure}

\begin{figure}[htbp]
	\centering
	{\includegraphics[width=\textwidth]{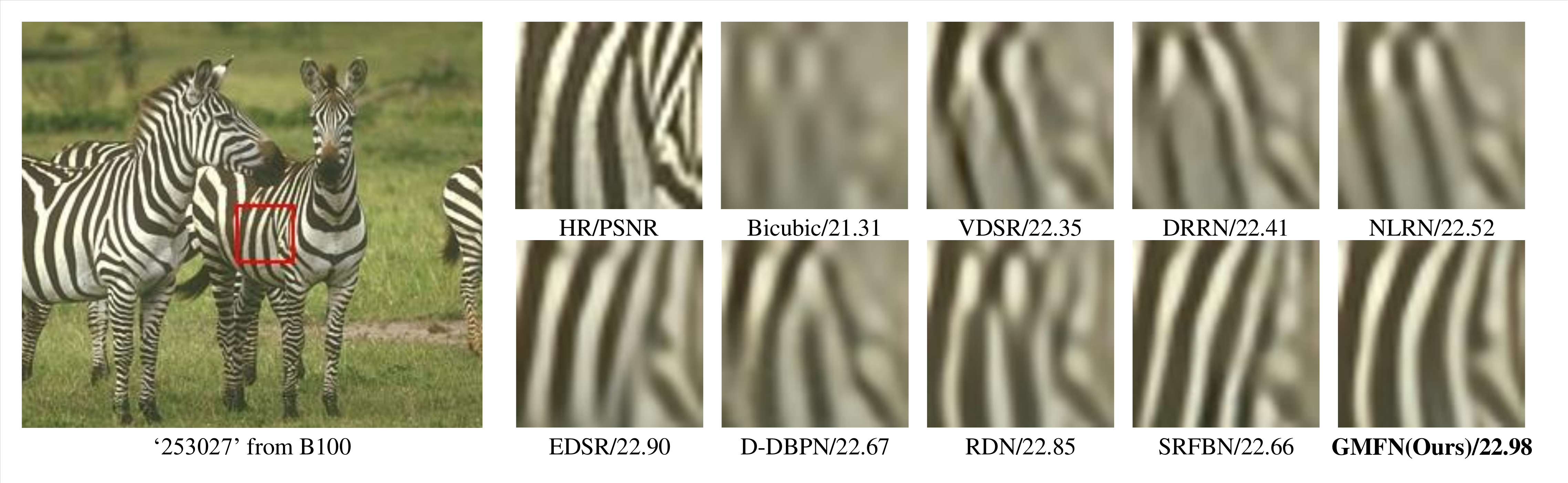}} 
	\caption{Qualitative results on `253027' with scale factor $\times$4. Only GMFN correctly reconstructs the direction of the zebra's stripes.}
	\label{fig:253027}
\end{figure}

\begin{figure}[htbp]
	\centering
	{\includegraphics[width=\textwidth]{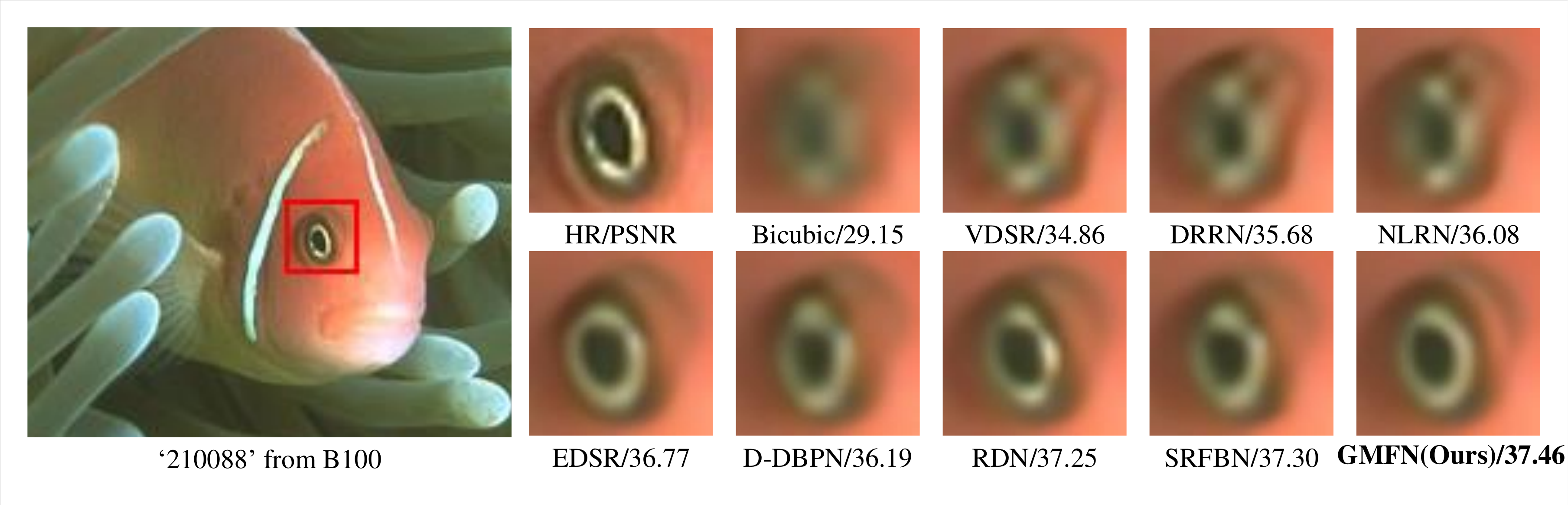}} 
	\caption{Qualitative results on `210088' with scale factor $\times$4. GMFN reconstructes a more vivid fisheye compared with other methods.}
	\label{fig:210088}
\end{figure}

\begin{figure}[htbp]
	\centering
	{\includegraphics[width=\textwidth]{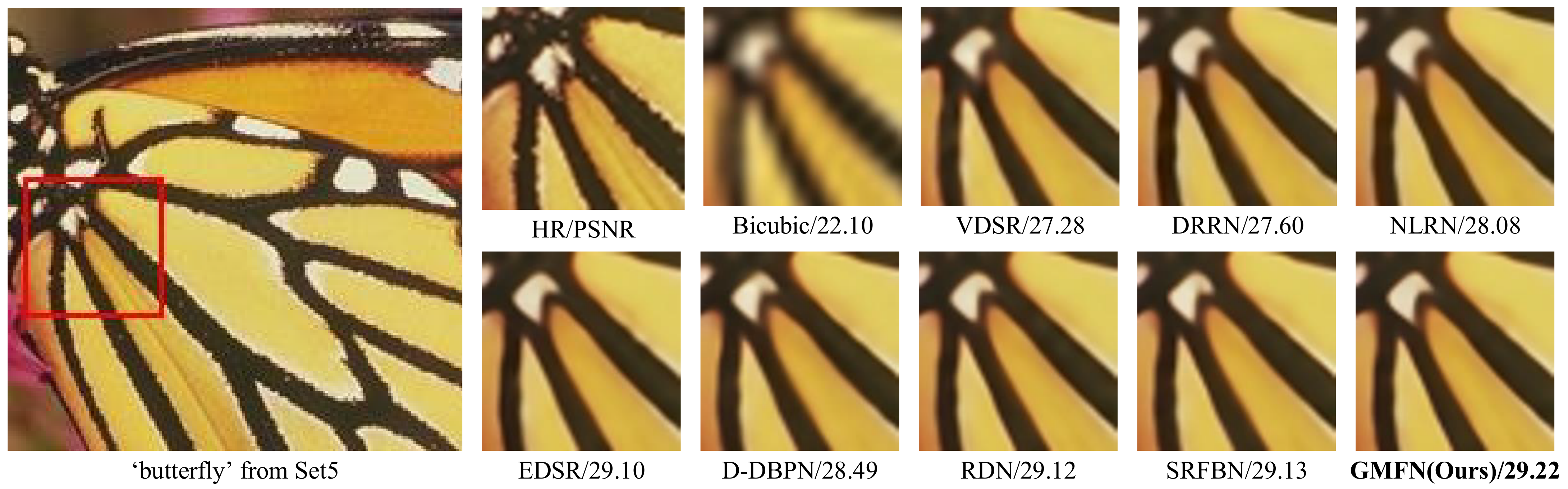}} 
	\caption{Qualitative results on `butterfly' with scale factor $\times$4.}
	\label{fig:butterfly}
\end{figure}

In Fig.~\ref{fig:img_044}-\ref{fig:butterfly}, we provide more qualitative results to prove the superiority of the proposed GMFN.

\end{document}